%% file: main.tex
\documentclass[option1,option2,etc.]{krantz}
\usepackage{fixltx2e,fix-cm}
\usepackage{amssymb}
\usepackage{amsmath}
\usepackage{graphicx}
\usepackage{subfigure}
\usepackage{makeidx}
\usepackage{multicol}
\usepackage{floatrow}
\usepackage[sorting=none]{biblatex}
\bibliography{bibtex_example}

\begin{document}
\title{Exploring Causes of Demographic Variations In Face Recognition Accuracy} 
\author{Gabriella Pangelinan, K.S. Krishnapriya, Vitor Albiero, Grace Bezold, Kai Zhang, Kushal Vangara, Michael C. King and Kevin W. Bowyer}

\maketitle

\input{Sections/Introduction}

\input{Sections/DemographicVariations}
\input{Sections/SkinTone}
\input{Sections/FaceSizeShape}

\input{Sections/BalancedTraining}
\input{Sections/BalancedPixels}
\input{Sections/Discussion}
\printbibliography
\end{document}

%% file: Sections/Introduction.tex
 \section{Introduction}
 \label{sec:intro}

\begin{figure}
\floatbox[{\capbeside\thisfloatsetup{capbesideposition={right,center},capbesidewidth=4cm}}]{figure}[\FBwidth]
{\caption{Apple's Face ID ad campaign touted ease of use and improved security for face recognition over the prior fingerprint standard, and made bold promises. %
}\label{fig:faceads}}
{\includegraphics[width=7cm]{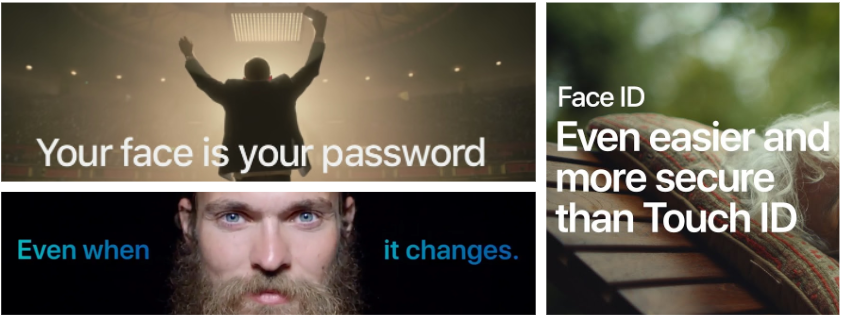}}
\end{figure}

Automated facial recognition (FR) technology dates back to the early 1970s, with Takeo Kanade's 1973 Ph.D. thesis \emph{Picture Processing System by Computer Complex and Recognition of Human Faces} \cite{Kanade} often cited as an early landmark work. However, it was not until the late 2010s that increased availability and power of FR technology increased its routine usage. In 2017, Apple introduced the iPhone X as the ``smartphone industry's benchmark'' \cite{dans_2018}, with their new facial identification 
system, Face ID, as a primary innovation and selling point \cite{faceid_ad1, faceid_ad2} (Figure \ref{fig:faceads}). The corresponding security guide \cite{applesupport} even claimed ``the probability that a random person in the population could look at your iPhone X and unlock it using Face ID is approximately 1 in 1,000,000 (versus 1 in 50,000 for Touch ID).''  Yet a security flaw was quickly publicized: some Chinese users reported that their iPhones opened for other, non-authorized individuals \cite{curtis_2017, zhao_2017}. The underpinnings of these incidents would soon be explored by the research community, leading to a major question: does face recognition perform equally across all demographics?

In 2018, Buolamwini and Gebru explored a related question in \textit{Gender Shades: Intersectional Accuracy Disparities in Commercial Gender Classification} \cite{buolamwini2018gender}.
Their work evaluated the accuracy of commercial gender classification software. 
For one classifier, they reported that error rates for lighter-skinned males were only 0.8\%, while error rates for darker-skinned females were dramatically higher, at 34.7\%. Provocative headlines like ``Facial Recognition Is Accurate, if You're a White Guy'' \cite{NYT} sparked public interest in Buolamwini and Gebru's work.
Media coverage generally failed to make any distinction between gender classification, the task of assigning a gender label to one face image, and face recognition, the task of deciding whether or not two face images are from the same person.

Government research organizations quickly addressed the growing public concern around possible ``bias'' in face recognition accuracy. %
As part of their 2018 biometric technology rally, the Department of Homeland Security assessed the effect of demographic factors on performance of commercial face biometric systems, as measured by transaction times and by similarity scores of pairs of images of the same person \cite{dhs}. In 2019, the National Institute of Standards and Technology (NIST) released a Face Recognition Vendor Test (FRVT) report on demographic accuracy variations in commercial face recognition tools \cite{nist_demographics}. The report provides extensive results across multiple datasets and matchers. However, NIST did not explore why the accuracy variations occur, and 
listed “analyze cause and effect” as an item under the heading “what we did not do” \cite{nist_demographics}.

\begin{figure}[!ht]
\centering
\includegraphics[width=\columnwidth]{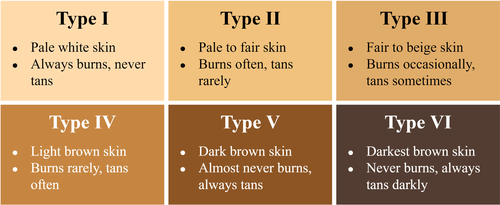}
\caption{The Fitzpatrick skin type scale \cite{fstscale} is used in dermatology. FST-inspired six-tone scales are often used in face recognition research.}
\label{fstScale}
\end{figure}

As public and private use of FR technology expands, it is more important than ever to assess how the technology interacts with people across gender and race. Current FR research relies on demographic labels that are either self-reported, annotated by manual reviewers, or generated by automated classifiers. 
Each possible source of demographic labels has weaknesses. Gender is typically treated as a binary male/female label, and race is typically treated in terms of a small number of discrete  categories (e.g. Caucasian, African American and Asian). 
Works focusing on skin tone often use a six-tone, light-to-dark scale like the Fitzpatrick skin type (FST) scale (shown in Figure \ref{fstScale}).
However, researchers have noted that true FST assignments are meant to be provided in person by a trained practitioner. Google AI's recent introduction of the ten-shade Monk Skin Tone (MST) scale \cite{monkscale} speaks to the growing desire, in industry and research alike, for greater inclusivity of the ``spectrum of skin tones we see in our society''  \cite{monkscale}.

The following sections highlight the ongoing collaboration between research teams at the Florida Institute of Technology and the University of Notre Dame to better understand how, and why, face recognition accuracy varies across demographics. Our work has provided insight into the complexity of this multi-faceted problem. There is still much work to be done - by researchers and developers alike - in the quest to create truly equitable and fair face recognition systems.
\newpage

%% file: Sections/DemographicVariations.tex
\section{Demographic Variations In Accuracy}
\label{sec:variations}

There is consensus in the research literature that face recognition accuracy is lower for females, who often have both a higher false match rate and a higher false non-match rate. In 2021, Albiero et al. \cite{vitorTIFS} presented the first experimental analysis to identify major causes of this ``gender gap'', which is quantifiable in terms of separation between female and male impostor and genuine distributions. Albiero et al. analyzed these distributions for Caucasian, African American and Asian demographics using two curated datasets and a top-performing open-source matcher (ArcFace \cite{arcface} trained on MS1Mv2 \cite{ms1m}). 

The Caucasian and African American demographics in \cite{vitorTIFS} come from the MORPH 3 dataset \cite{morph}, a large-scale collection of mugshot-style images acquired with controlled lighting and an 18\% gray background. Subjects generally have a frontal pose and neutral facial expression. Twins and mislabeled or duplicate images were removed from the curated version of MORPH, yielding the demographic distribution given in Table 0.1.

There is not yet a large generally-available dataset of Asian face images acquired in a controlled manner similar to that of MORPH.
For this reason, Albiero et al. curated a subset of the web-scraped Asian-Celeb dataset. Curation of a MORPH-comparable subset included removal of mislabeled or duplicate images and pose constraint, with the final composition given in Table 0.1.
\input{tables/morphnums.tex}

Albiero et al. \cite{vitorTIFS} used ArcFace, a state-of-the-art deep convolutional neural network (CNN), as the matcher, providing similarity scores between pairs of images. The experimental instance of ArcFace corresponds to a set of publicly-available weights, trained on the publicly-available MS1Mv2 dataset. ArcFace takes as input aligned face images resized to 112x112 pixels, extracts 512-d features, and matches using cosine similarity.
(In recent work, we have found that a newer instance of ArcFace trained on the Glint-360K dataset \cite{glint} gives higher overall accuracy than the version trained on MS1Mv2. However, the same demographic variations in accuracy are still seen with the Glint-trained ArcFace.)

\begin{figure}[!ht]
\centering
\includegraphics[width=\columnwidth]{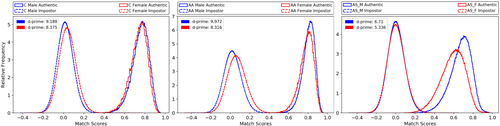}
\caption{ArcFace impostor and genuine (authentic) distributions for Caucasian, African American and Asian groups (from left to right). (Figure adapted from \cite{vitorTIFS}.)}
\label{arcfaceDist}
\end{figure}

Impostor and genuine distributions for each demographic group are given in Figure \ref{arcfaceDist}. The impostor distribution contains match scores resulting from comparisons between images of two different individuals, or \emph{non-mated} image pairs. The genuine distribution's match scores result from comparisons between different images of the same individual, or a \emph{mated} pair. 

For all three demographic groups, the female impostor distribution is shifted toward higher similarity scores than its male counterpart. This shift indicates a higher false match rate (FMR): that is, a higher rate at which a given image instance is classified as an incorrect identity (a ``false accept''). Each female genuine distribution ranges over lower similarity scores, indicating a higher false non-match rate (FNMR): a higher rate at which a given image instance is not correctly classified with its true identity (a ``false reject''). For each group, the d-prime value - a statistical measure of a recognition system's ability to distinguish between samples - is lower for females than males. The lower d-prime value indicates that the matcher is less effective in differentiating between images of females than males.

\begin{figure}[!ht]
\centering
\includegraphics[width=\columnwidth]{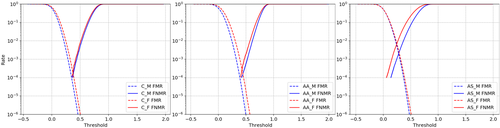}
\caption{ArcFace FMR and FNMR for Caucasian, African American and Asian groups (from left to right). (Figure adapted from \cite{vitorTIFS}.)}
\label{arcfaceFMRs}
\end{figure}

The FMR and FNMR curves for each group are given in Figure \ref{arcfaceFMRs}. The FNMR is consistently higher for females than males across all datasets, though the corresponding gender gap is larger for Asian-Celeb and MOPRH African American groups than MORPH Caucasian.  For both MORPH datasets, FMR is significantly higher for females than males; for Asian-Celeb, the cross-gender FMRs are more similar. Albiero et al. speculate that this difference can be attributed to the fact that Asian-Celeb is a web-scraped dataset - despite curation, impostor pairs may still have factors other than gender affecting the FMR.

With this foundational understanding of the gender gap, we proceed to review analyses of its various speculated causes. Namely, we explore how cross-demographic recognition performance varies with skin tone, face geometry, representation in training data, and face pixel information.
\newpage

%% file: tables/morphnums.tex
\begin{table}[h]
\begin{tabular}{l|cc|cc|}
\cline{2-5}
 & \multicolumn{2}{c|}{\textbf{Female}} & \multicolumn{2}{c|}{\textbf{Male}} \\ \cline{2-5} 
 & \multicolumn{1}{c|}{Images} & Subjects & \multicolumn{1}{c|}{Images} & Subjects \\ \hline
\multicolumn{1}{|l|}{\textbf{Caucasian}} & \multicolumn{1}{c|}{10,941} & 2,798 & \multicolumn{1}{c|}{35,276} & 8,835 \\ \hline
\multicolumn{1}{|l|}{\textbf{African American}} & \multicolumn{1}{c|}{24,857} & 5,929 & \multicolumn{1}{c|}{56,245} & 8,839 \\ \hline
\multicolumn{1}{|l|}{\textbf{Asian}} & \multicolumn{1}{c|}{43,356} & 6,083 & \multicolumn{1}{c|}{73,376} & 12,673 \\ \hline
\end{tabular}
\label{tab:morphnums}
\caption{Number of images and subjects in the curated datasets used in \cite{vitorTIFS}. (Table adapted from \cite{vitorTIFS}.)}
\end{table}

%% file: Sections/SkinTone.tex
\section{Does Darker Skin Tone Cause Increased False Match Rate?}
\label{sec:skintone}

Media coverage has hyped darker skin tone as a primary driver of recognition accuracy varying across race. Articles in prominent news sources have stated that ``face recognition tech is less accurate the darker your skin tone'' (BBC) \cite{BBC} and ``the darker the skin tone, the more errors arise'' (New York Times) \cite{NYT}. 

Research studies also seem to have supported this notion. Cook et al. \cite{dhs} reported that darker skin tone is associated with lower similarity scores for mated pairs. Grother \cite{Grother_2017} found that ``African Americans give slightly lower FNMR than Whites'' and a ``much higher FMR''. Wang et al. \cite{wang2018} reported that Caucasians have higher verification accuracy measured on in-the-wild imagery with intentionally difficult pairs. Grother et al. \cite{nist_demographics} observed that Caucasians have more false negatives in mugshot-quality images, while African Americans have more false negatives in lower-quality images. Interestingly, Howard et al. \cite{howard2019} found that for males younger than 40, African American FMR is greater than Caucasian, with the opposite effect observed for males aged 40 or older.

Table \ref{prevFR} provides an overview of previous results regarding African American versus Caucasian FR accuracy. (We exclude analyses based on ROC curves, as they can hide the fact that different demographics achieve the same FMR at different similarity thresholds.)
The general consensus is that, for a fixed score threshold, African Americans experience a higher FMR and Caucasians experience a higher FNMR.
Many different matchers have been tested across the highlighted works.
The only dataset that is generally available to researchers and contains mugshot-style images is MORPH.
The only other dataset that has been used and is generally  available to researchers is IJB-C \cite{ijbc}.
The datasets that are available are typically not balanced on skin tone, race or other factors.

\begin{table}[!ht]
  \centering %
  \caption{Previous comparisons of African American (AA) vs. Caucasian (C) accuracy. (Table adapted from \cite{kksIssues}.)}\label{prevFR}
  \includegraphics[width=\textwidth]{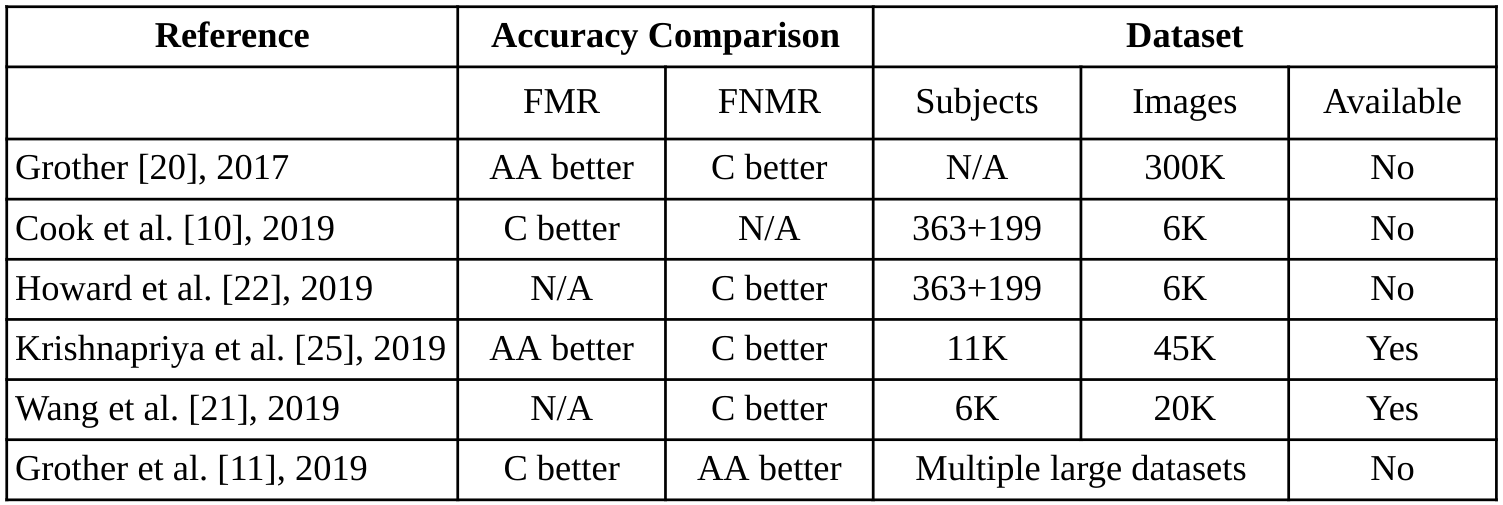}
\end{table}

\subsection{Skin Tone and False Match Pairs}

As concern around possible racial bias in face recognition continued to grow, Krishnapriya et al. \cite{kks2019} observed that existing comparisons across African American and Caucasian images had confounding effects of skin tone and face morphology. It was impossible to ascertain which of the two effects was the driving factor.
(In light of the recent work of Albiero \cite{vitorTIFS}, we might now add that differing social conventions for hairstyle present another confounding effect.)
In \textit{Issues Related to Face Recognition Accuracy Varying Based on Race and Skin Tone}, Krishnapriya et al. \cite{kksIssues} published the first experiment intended to isolate the effect of skin tone alone on accuracy. To directly test the premise that face recognition is less accurate for darker skin tones, they examined a range of tones within the single demographic of African American male (AAM) using the MORPH dataset.

They used two matchers to produce AAM impostor distributions: ArcFace (as described in Section \ref{sec:variations}) and a publicly-available VGGFace2 model \cite{keras_vggface}. VGGFace2 is representative of the state-of-the-art in CNN matchers prior to ArcFace. It is based on the popular ResNet-50 network structure and trained on the VGGFace2 dataset \cite{vggface} with standard softmax loss. Faces are detected, aligned, and resized to 224×224 pixels, and a 2048-d feature vector is taken from the next-to-last layer. As with ArcFace, cosine similarity is measured between feature vectors. 

\begin{figure}[!ht]
\centering
\includegraphics[width=\columnwidth]{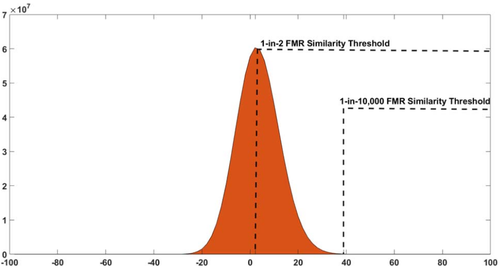}
\caption{The ArcFace impostor distribution for the African American male cohort of MORPH, with the center and HST indicated. (Figure adapted from \cite{kksIssues}.)}
\label{aamDistKKS}
\end{figure}

Krishnapriya et al. compared the frequency of images with darker skin tone in two regions of the AAM impostor distribution.
The high-similarity tail (HST) is the region containing the non-mated image pairs that are most likely to cause false matches.
To represent the HST, image pairs were sampled from just above the commonly used 1-in-10,000 (1-in-10k) FMR threshold.
This threshold corresponds to the non-mated similarity score at which one error occurs for 10,000 image-pair comparisons.
The center of the AAM impostor distribution was sampled as the ``no-false-match region''. Image pairs are selected from just above the 1-in-2 FMR threshold.
The two regions are visualized in Figure \ref{aamDistKKS}.

\begin{figure}[!ht]
    \centering
    \includegraphics[width=0.7\textwidth]{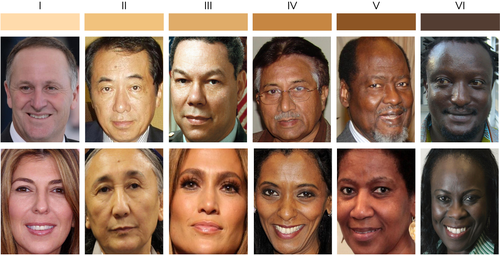}\hfill
    \caption{Exemplar images of skin tone ratings I-VI used by manual raters for reference. (Figure adapted from \cite{krishnapriya2022analysis}.)}
    \label{fig:Exemplar_Img}
\end{figure}

Each image was assigned a skin tone rating from I (lightest) to VI (darkest), inspired by the Fitzpatrick skin type scale, by three independent observers. The observers were shown exemplar images (Figure \ref{fig:Exemplar_Img}) selected from the well-known and manually annotated IJB-C dataset \cite{ijbc} to guide their ratings. If two or three of the raters agreed on a given skin tone rating, that rating was used. If all three gave different ratings, the middle rating of the three was used. Across the four total samplings (two samples from each of the two matchers), two or three viewers agreed on the skin tone rating for 93\% to 96\% of images. 

\begin{figure}[!ht]
\centering
\includegraphics[width=\columnwidth]{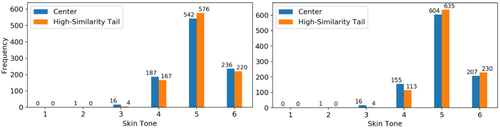}
\caption{Comparison of skin tone ratings from the center and HST of the AAM impostor distribution from ArcFace (left) and VGGFace2 (right). (Figure adapted from \cite{kksIssues}.)}
\label{fig:3Ratings}
\end{figure}

The two sets of manual ratings are visualized in Figure \ref{fig:3Ratings}. If darker skin tone \emph{is} a cause of increased false matches, we would expect to see a significantly higher frequency of skin tones IV-VI in HST images. However, for both matchers, the distribution of skin tone ratings varies only slightly between center and HST images. For both sets of images, the most frequent skin tone rating is V, followed by IV, then VI, then III, with almost no images rated I or II. There is no clear shift toward higher or lower skin tone ratings in either region of the impostor distribution.

\subsection{Standardized Skin Tone and False Match Pairs}

\begin{figure}[h]
\floatbox[{\capbeside\thisfloatsetup{capbesideposition={right,center},capbesidewidth=4cm}}]{figure}[\FBwidth]
{\caption{Noticeable variations exist in the gray background of the original MORPH images (top row), while their color-corrected versions (bottom) more accurately display the standard 18\% gray background.}\label{fig:wacvCC}}
{\includegraphics[width=7cm]{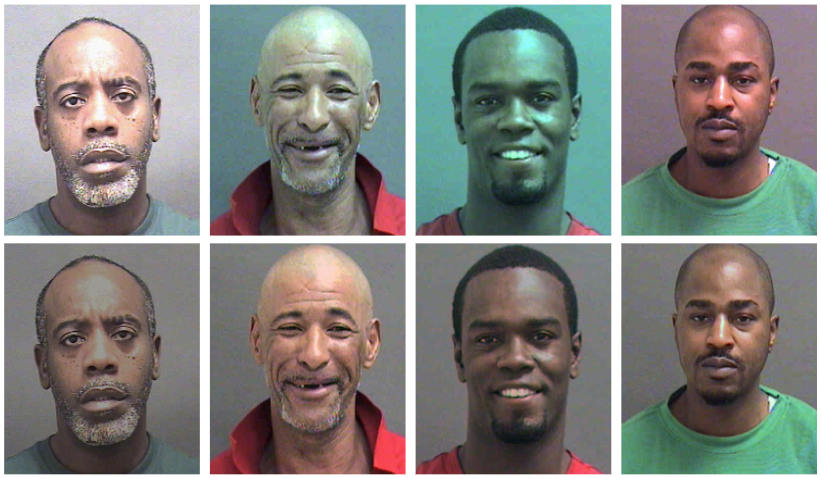}}
\end{figure}

Extending the aforementioned work, Krishnapriya et al. conducted a second experiment \cite{krishnapriya2022analysis}, adding color correction, an automated skin tone rating system, and additional manual raters. In agreement with Howard et al. \cite{howard2021}, they determined that a color correction step is crucial to the accurate estimation of apparent skin tone, to account for inconsistent image lighting or coloration. The controlled capture environment of the MORPH dataset allows for color correction of the images based on the standard 18\% gray background. The original face images in the top row of Figure \ref{fig:wacvCC} (adapted from \cite{krishnapriya2022analysis}) feature a notable range of variation in the background color. After color correction, the images show a more consistent gray background.

\begin{figure}[!ht]
\centering
\includegraphics[width=.6\columnwidth]{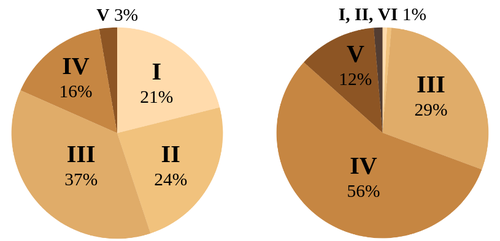} \vspace{.2cm}\\
\includegraphics[width=.6\columnwidth]{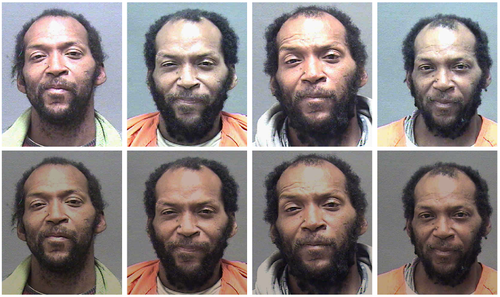}
\caption{Automated ratings assigned to a single subject, before and after color correction.}
\label{fig:skinTypeEx}
\end{figure}

 Even a \emph{single} subject's skin tone may vary significantly, both by visual and automated estimation, across their image set. Figure \ref{fig:skinTypeEx} gives an example of the automated skin type ratings associated with 150 face image instances of a single subject. The two pie charts give the distribution of skin type ratings based on the subject's original, uncorrected images (left) and the color-corrected versions (right). Four sample images are shown before and after correction in the top and bottom rows, respectively.
 
 For an African American subject, we expect higher-valued skin type ratings. However, the ratings of the original images in Figure \ref{fig:skinTypeEx} primarily span over I-III (lighter) values. Inspection reveals overexposure: spectral highlights are present across the subject's forehead, nose and cheek regions, interpreted as artificially-lightened pixels. The majority of the corrected image ratings range from III-V, which is likely a more accurate estimation of the subject's true skin tone.

After color correction, manual and automated skin tone ratings were recorded for the center and HST images. Six manual raters independently assigned Fitzpatrick-inspired skin tone ratings from I to VI. For each image, the average of the six ratings was used as its consensus manual rating. Automated ratings were generated using six-toned, light-to-dark individual typology angle (ITA) measurements, which have been used in previous research \cite{del2006relationship,diversityinfaces} to directly assess skin tone from imagea. Krishnapriya et al. produced automated ratings for each image by mapping ITA values onto the FST-like scale.

\begin{figure}[!ht]
\centering
\includegraphics[width=.85\columnwidth]{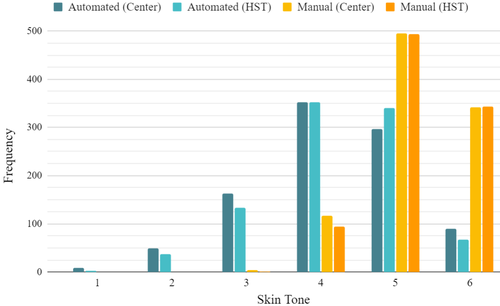}\hfill
\caption{Distribution of ratings on color-corrected center versus HST images. (Figure adapted from \cite{krishnapriya2022analysis}.)}
\label{fig:centVShst}
\end{figure}

Figure \ref{fig:centVShst} gives the distribution of automated and consensus manual skin tone ratings for color-corrected images from the center and high-similarity tail of the African American male impostor distribution. The two rating sets are generally consistent across center and HST images, reinforcing the previous finding that region of impostor distribution has little effect on skin tone rating.

\subsection{Summary of Skin Tone Results}

In their two experiments designed to directly assess the impact of skin tone on FMR, Krishnapriya et al. find no clear evidence to support the idea that skin tone is a driving factor behind the higher false match rate typically reported for African American individuals.
That is, their results do not support a general conclusion that darker skin tone, in and of itself, causes an increased FMR. 
They note that the premise “face recognition is less accurate for darker skin tones” appears to be an over-simplification of previous research results on the ROC curves or FMRs for African Americans versus Caucasians.

Though the same FMR is typically achieved for different demographics at different thresholds, real-world recognition systems often use the same threshold across demographics. This practice has likely contributed to the common conclusion that the higher FMR for darker-skinned individuals is driven by skin tone. Krishnapriya et al. conclude that while the issue certainly merits further study, their results suggest that factors \emph{other} than skin tone should be considered in looking for the cause of race-asymmetric recognition accuracy.

\newpage

%% file: Sections/FaceSizeShape.tex
\section{Do Gender Differences In Face Size and Shape Cause Accuracy Differences?}
\label{sec:face_size_shape}

The study of facial landmark points  by Farkas et al. \cite{Farkas2005} is well known in the face recognition research community.
Demographic differences in face size and shape have been studied in other fields as well. For example, these factors impact the design of personal protective equipment \cite{Zhuang2010}.
In this section, we investigate several 
  female/male differences in face size and shape to assess their impact on the gender gap. To collect face measurements, we use a 3D+2D face image dataset originally collected at the University of Notre Dame and distributed as part of the Face Recognition Grand Challenge (FRGC) program \cite{frgc}.

\begin{table}[!ht]
\begin{tabular}{|c|cc|cc|c|}
\hline
\multicolumn{1}{|l|}{} & \multicolumn{2}{c|}{Male} & \multicolumn{2}{c|}{Female} & Male - Female \\ \hline
Distance & \multicolumn{1}{c|}{Average} & SD & \multicolumn{1}{c|}{Average} & SD & SD \\ \hline
n-gn & \multicolumn{1}{c|}{121.3} & 6.8 & \multicolumn{1}{c|}{111.8} & 5.2 & +1.6 \\ \hline
sn-gn & \multicolumn{1}{c|}{71.9} & 6 & \multicolumn{1}{c|}{65.5} & 4.5 & +1.5 \\ \hline
n-sn & \multicolumn{1}{c|}{53} & 3.5 & \multicolumn{1}{c|}{48.9} & 2.6 & +0.9 \\ \hline
ex-ex & \multicolumn{1}{c|}{89.4} & 3.6 & \multicolumn{1}{c|}{86.8} & 4 & -0.4 \\ \hline
\end{tabular}
\caption{Select mean and standard deviation (SD) values and male-female SD differences, as reported in \cite{Farkas2005}.}
\label{farkas_Table1}
\end{table}

Table \ref{farkas_Table1} from \cite{Farkas2005} gives average face size and shape measurements for individuals representing a “North American White Young Adult Population”. We are particularly interested in two measurements. The \textbf{ex-ex} distance is measured between the outer eye corners (``exocanthion'' points) and relates to face width. The average \textbf{ex-ex} distance is 89.4 mm for males and 86.8 mm for females. The \textbf{n-gn} distance relates to face height. It is measured from the point of high curvature on the centerline of the nasal bridge (“nasion”, or {\bf n}) to the point on the midline of the chin where face curvature turns (“gnathion”, or {\bf gn}). The average \textbf{n-gn} distance is 121.3 mm for males and 111.8 mm for females.

Table \ref{farkas_Table1} also gives the standard deviation values for each distance measurement.  The standard deviation of {\bf ex-ex} distance is 3.6 mm for males and 4 mm for females. The standard deviation of {\bf n-gn} distance is 6.8 mm for males and 5.2 mm for females. These differences in face width and height and their standard deviations suggest that there is a gender-based difference in the average aspect ratio of the face.  

In this section, we examine whether the female/male differences in face measurements impact the gender gap in recognition accuracy. We design experiments to assess the following factors:
\begin{itemize}
    \item \textit{Face width:} Does the larger average {\bf ex-ex} distance for males result in a greater amount of face information and better accuracy?
    \item \textit{Variation in face height:} Does the larger standard deviation in {\bf n-gn} distance for males lead to lower similarity scores for impostor (non-mated) pairs of images?
    \item \textit{Face aspect ratio:} Does balancing the aspect ratio of male and female faces decrease the gender gap?
\end{itemize}

\subsection{Dataset, Landmark Points and Distances}

\begin{figure}[!ht]
  \centering
  \includegraphics[width=0.45\linewidth]{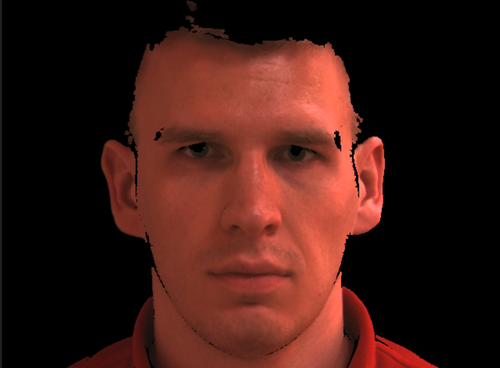}
  \includegraphics[width=0.45\linewidth]{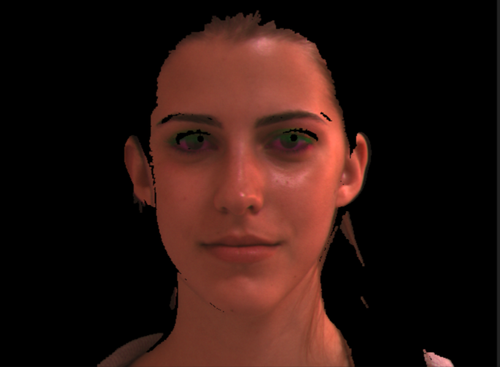}
  
 \caption{Example 2D RGB face images acquired with the Minolta Vivid 910 3D scanner. Pixels are mapped to black if there is no corresponding (x,y,z) location.}
\label{fig:Minolta_faces}
\end{figure}

The Minolta Vivid 900/910 sensor captures a series of 2D color images (480x640 pixels) and uses triangulation to obtain 3D (x,y,z) coordinates. The resulting data points are registered, associating pixels in the 2D image with their 3D (x,y,z) locations in the scene.
Sample 2D images from the FRGC dataset are shown in Figure~\ref{fig:Minolta_faces}. The 3D data points are missing in areas where the face surface is too far from the sensor, or where the projected light beam is not seen clearly in the 2D image, e.g. the pupils and some hair regions. The pixels shown as black in Figure~\ref{fig:Minolta_faces} have no corresponding 3D point.

\begin{figure}[!ht]
  \centering
  \includegraphics[width=0.70\linewidth]{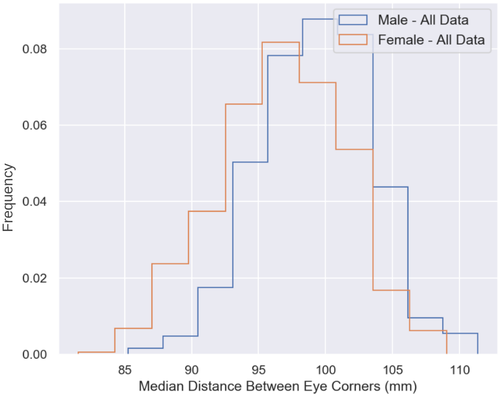}
 \caption{Distribution of female/male face width ({\bf ex-ex}) values.}
\label{fig:exocanthion_distance_distributions}
\end{figure}

We begin by measuring average face width. Two independent viewers mark {\bf ex} points in the 2D images. The corresponding 3D coordinates are used to compute {\bf ex-ex} distance. Images with more than 0.5 cm difference in the independently-computed {\bf ex-ex} distance are marked again. If this difference persists in a second round of marking, the image is dropped from analysis. %
From the final set of {\bf ex-ex} distances, the median value is taken as the size of the face for a given image. The two rounds of {\bf ex-ex} markings resulted in a dataset of 1618 images representing 583 females and 480 males. The resulting distribution of face width is compared for females and males in Figure~\ref{fig:exocanthion_distance_distributions}.

\begin{figure}[!ht]
  \centering
  \includegraphics[width=.8\linewidth]{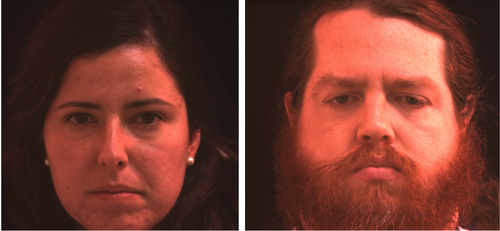}
 \caption{Example problematic images for marking face height landmark points. \textit{Left:} The {\bf gn} point is cut off at the bottom of the image. \textit{Right:} The {\bf gn} point is obscured by facial hair.}
\label{fig:faces_dropped}
\end{figure}

\begin{figure}[!ht]
  \centering
  \includegraphics[width=0.9\linewidth]{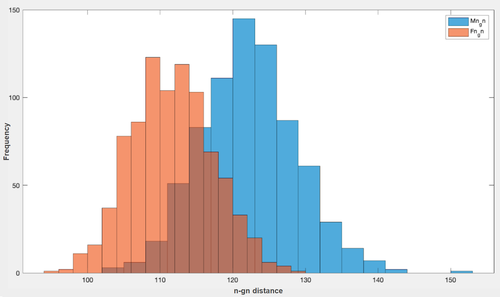}
 \caption{Distributions of {\bf n-gn} distances for females and males.}
\label{fig:original_n-gn_measurements}
\end{figure}

We follow the same process to measure face height variation, marking {\bf n} and {\bf gn} points on each image and calculating {\bf n-gn} distances. In some images, {\bf gn} landmark points could not be accurately marked due to issues like facial hair or the chin being cropped. Sample removed images are shown in Figure~\ref{fig:faces_dropped}. The distributions of {\bf n-gn} distances for females and males are given in Figure~\ref{fig:original_n-gn_measurements}. The means and standard deviations of the {\bf n-gn} distances for females and males, given in Table~\ref{tab:landmark_distances2}, are similar to those reported by Farkas \cite{Farkas2005}.  

\begin{table}[]
    \centering
    \begin{tabular}{|c||c|c||c|c|}
    \hline
       distance & male (SD) & female (SD) & Farkas male & Farkas female \\ \hline
        \hline
       n-gn   & 122.3 (6.6) & 111.6 (5.6) & 121.3 (6.8) & 111.8 (5.2) \\ \hline
      sn-gn   &  70   (5.6) & 63.1 (4.8)  & 71.9 (6) & 65.5 (4.5) \\ \hline
      n-sn    &  54.3 (3.6) & 50.1 (3.3)  & 53 (3.5) & 48.9 (2.6) \\ \hline
    \end{tabular}
    \caption{Measured distances for landmark points (left), compared to those reported in \cite{Farkas2005} (right).}
    \label{tab:landmark_distances2}
\end{table}

\begin{figure}[!ht]
  \centering
  \includegraphics[width=0.9\linewidth]{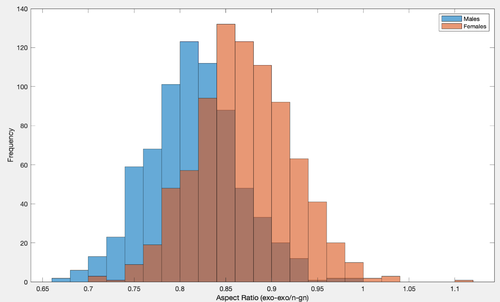}
 \caption{Distributions of female and male aspect ratios.}
 \label{fig:female_male_aspect_ratios}
\end{figure}

Finally, we compute face aspect ratio, dividing face width ({\bf ex-ex} distance) by height ({\bf n-gn}). The aspect ratio indicates whether overall face shape is more elongated or more rounded. Figure~\ref{fig:female_male_aspect_ratios} shows the distributions for the female and male aspect ratio measurements. Females tend to have rounder faces, while males tend to have more elongated faces.

\subsection{Do Size and Shape Differences Drive Accuracy Differences?}

\begin{figure}[!ht]
  \centering
  \includegraphics[width=0.45\linewidth]{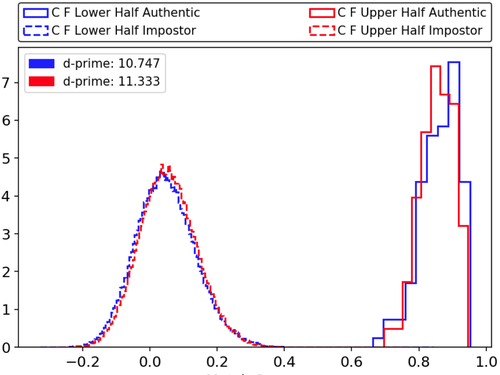}
  \includegraphics[width=0.45\linewidth]{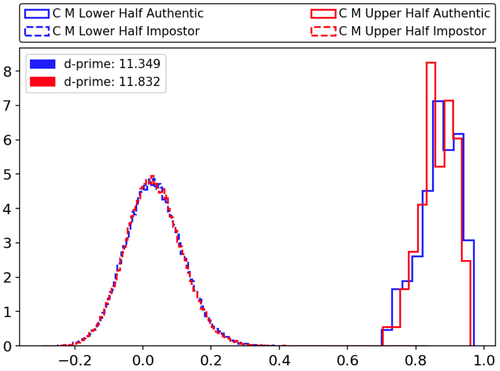}
  \\
 \caption{Impostor and genuine (authentic) distributions of face width groups, with the Caucasian female plot on the left and male on the right.}
    \label{fig:impostor_genuing_face_size_female}
\end{figure}

Our first experiment analyzes the impact of face width. We divide both male and female 2D images into two {\bf ex-ex} distance groups. Faces in the first group of images (``Lower Half'') have smaller-than-median {\bf ex-ex} distance. Faces in the second group (``Upper Half'') have larger-than-median {\bf ex-ex} distance. We then use ArcFace to generate impostor and genuine distributions for each group and gender. These distributions are shown in Figure ~\ref{fig:impostor_genuing_face_size_female}. 

For both genders and distribution types, there is no clear difference in lower- and upper-half scores. In both plots of Figure \ref{fig:impostor_genuing_face_size_female}, there is significant overlap between the blue and red curves (corresponding to the lower- and upper-half groups, respectively). Thus, we find no evidence that differences in average face width contribute to the observed differences in female and male impostor or genuine distributions.

The next experiment considers face height variation. We are interested in whether faces with greater height variation yield lower non-mated scores. Lower non-mated scores correspond to an improved impostor distribution and imply a lower false match rate. Given that males have a larger average {\bf n-gn} distance and standard deviation (Table \ref{tab:landmark_distances2}), we design an experiment to balance {\bf n-gn} distance across male and female face images. 

We create subsets of male and female images that have a nearly equal standard deviation in the distribution of {\bf n-gn} distances using an iterative two-step process. In the first step, we remove the two male face images with the maximum and minimum {\bf n-gn} distances. Then, the standard deviation of the remaining male set is compared to the female standard deviation (5.6 mm). After 21 iterations, the remaining subset of male images had an {\bf n-gn} standard deviation of 5.9 mm. While this balancing approach does not perfectly balance the overall distributions, we find it sufficient to test the impact of the {\bf n-gn} distribution on the impostor distribution.

\begin{figure}[!ht]
  \centering
  \includegraphics[width=0.9\linewidth]{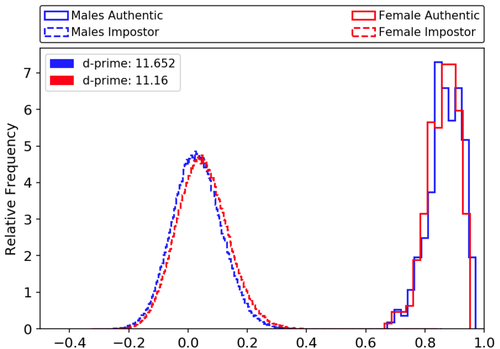}
 \caption{Impostor and genuine distributions of the male and female subsets balanced on n-gn standard deviation (height).}
 \label{fig:sd_balanced_impostors}
\end{figure}
The resulting impostor and genuine distributions are shown in Figure~\ref{fig:sd_balanced_impostors}. Balancing the standard deviation of the female and male {\bf n-gn} distances has no noticeable effect on the difference between the female and male impostor distributions. The female impostor distribution (indicated by the red dashed line) remains skewed toward higher similarity scores than the male distribution. That is, we find no clear evidence that controlling for face height decreases the gender gap seen in impostor distributions.

\begin{figure}[!ht]
  \centering
  \includegraphics[width=0.9\linewidth]{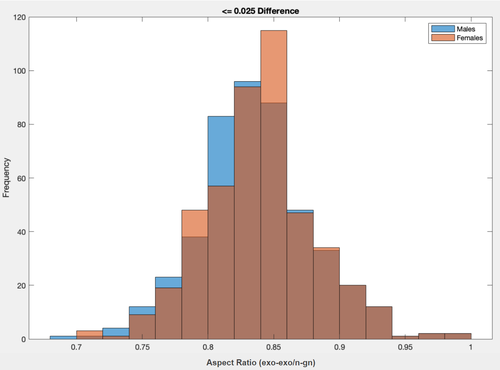}
 \caption{Balanced female/male aspect ratio distributions.}
 \label{fig:balanced_aspect_ratios}
\end{figure}

As a final experiment, we combine the previous width and height measurements to study cross-gender aspect ratio. We begin by balancing female/male images with respect to face aspect ratio. Each male image is paired with a female image that has an aspect ratio difference less than or equal to 0.025. This balancing yielded a set of 464 male images and 464 female images with approximately-matched aspect ratios. The resultant image distributions are given in Figure~\ref{fig:balanced_aspect_ratios}. 

\begin{figure}[!ht]
  \centering
  \includegraphics[width=0.9\linewidth]{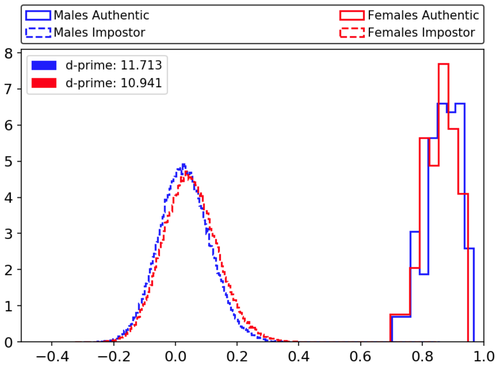}
 \caption{Impostor and genuine (authentic) distributions for female/male image sets balanced on aspect ratio.}
 \label{fig:balanced_aspect_impostors}
\end{figure}

The impostor and genuine distributions for the subsets balanced on aspect ratio are given in Figure \ref{fig:balanced_aspect_impostors}. Again, we find no observable impact on the distributions. The female distributions remain closer together than male distributions, with impostor scores skewed toward higher values and genuine scores toward lower values.

\subsection{Summary of Face Size and Shape Results}

In this section's experiments, we measure facial landmark points and distances using 3D+2D face images. Our measurements and results align with those previously reported by Farkas \cite{Farkas2005}. The measurements reveal consistent and notable differences in face size and shape for males versus females. In general, males have larger average values for facial measurements. Their faces are usually longer and have larger aspect ratios. Additionally, men tend to have larger standard deviations in face measurements than women, a hypothesized reason why male faces may be easier to differentiate than female faces.

In the three experiments, we analyze how facial width, height variation, and aspect ratio impact male and female impostor and genuine score distributions. In each case, balancing female/male images with respect to the given measurement has no significant effect on the resulting distributions. We thus find no clear evidence that the observed facial differences contribute to the gender gap.
\clearpage
\newpage

%% file: Sections/BalancedTraining.tex
\section{Does Balanced Training Data Equalize Accuracy In Test Data?}
\label{sec:balancedtraining}

Some authors have suggested that demographic differences in accuracy are due to imbalanced training datasets \cite{Karkkainen2019}, which tend to have ``a higher number of face images from [the] Caucasian ethnicity and males'' \cite{Vera-Rodriguez2019}. Researchers note that biases resulting from such imbalances do not seem to be intentional \cite{Drozdowski2020}, but assume that systems built on skewed data are simply ``bound to produce biased models'' \cite{diversityinfaces}.  The speculation from this line of thought is that explicitly gender-balanced training data should eliminate or at least greatly reduce the gender gap.

\subsection{Training with Balanced Data}

Albiero et al. first addressed the issue in \textit{Analysis of Gender Inequality In Face Recognition Accuracy} \cite{vitorkks_gender}. For an accurate assessment of how balancing training data impacts recognition, they trained a network from scratch. Using the ResNet-50 backbone, they trained two separate networks with combined margin loss on subsets of the VGGFace2 \cite{vggface} and MS1Mv2 datasets. These subsets were explicitly balanced on number of male and female persons and images. 

\begin{figure}[!ht]
\centering
\includegraphics[width=\columnwidth]{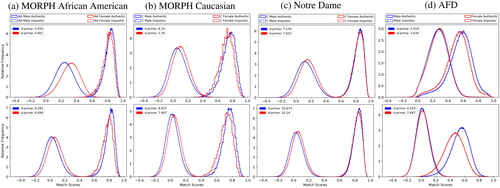}
\caption{Male and female distributions using networks trained on the gender-balanced VGGFace2 (top) and MS1Mv2 (bottom) datasets. (Figure adapted from \cite{vitorkks_gender}.)}
\label{balancedDists}
\end{figure}

Results for each model are given in Figure \ref{balancedDists}, tested on four datasets: MORPH's African-American and Caucasian cohorts, the Notre Dame dataset \cite{frgc}, and the Asian Faces Dataset (AFD) \cite{asianfacesdataset}, curated as described in Section \ref{sec:variations}.

The MS1Mv2-trained model achieves the best distributions across all datasets and both genders. Both models give a decreased d-prime difference between males and females in the Notre Dame dataset versus initial ArcFace results. However, all results in Figure \ref{balancedDists} still feature female impostor and genuine distributions that are closer together than the male distributions.

\subsection{Training with Skewed Data}

With the previous result in mind, they broadened their analysis to include both balanced and methodically {\bf imbalanced} data in \textit{How Does Gender Balance in Training Data Affect Face Recognition Accuracy?} \cite{vitor_gender}. For both the VGGFace2 and MS1Mv2 datasets, they constructed seven training subsets with images explicitly selected to vary gender representation. The descriptions of these subsets, and a visualization adapted from \cite{vitor_gender}, are given below.
\begin{itemize}
    \item \emph{Full}: full dataset with no modifications
    \item \emph{Balanced}: male images removed to equal the number of female images 
    \item \emph{F100}: 100\% female images
    \item \emph{M25F75}: 75\% female, 25\% male images
    \item \emph{M50F50}: 50\% female, 50\% male images
    \item \emph{M75F25}: 25\% female, 75\% male images
    \item \emph{M100}: 100\% male images
\end{itemize}

\begin{figure}[!ht]
\centering
\includegraphics[width=.6\columnwidth]{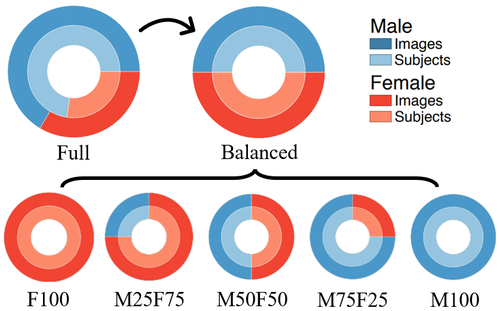}
\label{gbApproach}
\end{figure}

The five latter subsets consist of images randomly selected from the Balanced subset. Final counts of subjects and images in each training subset are given in Figure \ref{ratioTable}. They continued to use MORPH's Caucasian and African-American cohorts and the Notre Dame dataset for testing.

\input{tables/ratioTable.tex}

They again selected the ResNet-50 architecture to train on the subsets, with three different loss functions: the standard softmax loss, the newer combined margin loss, and the non-classification triplet loss. First, they analyzed male, female, and average accuracy on the Full dataset and Balanced subset. Results for each of the testing datasets are given in Table \ref{full_balanced}. 

\input{tables/full_balanced.tex}

Interestingly, with a softmax loss function, both Balanced subsets give the best male, female, and average accuracy versus Full. Combined margin loss gives the opposite result: best cross-gender and average accuracy is generally achieved with the Full subsets. Triplet loss gives inconsistent results across the two subsets. In general, Albiero et al. find that the MS1Mv2 training dataset and combined margin loss yield the highest overall accuracy. 

Selecting the best training set (MS1Mv2) and loss function (combined margin), they proceeded to analyze the five remaining subsets. They expected the best single-gender accuracy when training with M100 for males and F100 for females. They expected higher average accuracy on the balanced (M50F50) subset. The results are reported in Table \ref{mixedTable}. Accuracy value are colored from green to red, where green indicates the best result is achieved with the expected training and red the opposite.

\begin{table}[!ht]
  \centering %
  \caption{Gender accuracy (\%) with different balancing proportions. All trainings have the same number of subjects and images. (Table adapted from \cite{vitor_gender}.)}\label{mixedTable}
  \includegraphics[width=.9\textwidth]{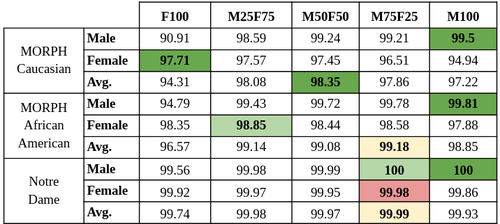}
\end{table}

For the MORPH Caucasian cohort, each best accuracy is achieved with the expected training. The African-American cohort, however, achieves the best female accuracy on the 25\% male training subset, and best average accuracy on the 75\% male subset. The 75\% male subset also gives the best overall accuracy for the Notre Dame dataset. 

\subsection{Summary of Balanced Training Results}

In their first study \cite{vitorkks_gender} on gender-balancing the training data, Albiero et al. found that an equal number of male and female subjects and images does \emph{not} improve the separation between the female impostor and genuine distributions. Instead, \cite{vitor_gender} showed that, when training set and loss function are both selected to maximize accuracy, \emph{imbalanced}, male-dominated training data results in higher female, male, and average accuracy. They propose that, with a good combination of training set and loss function, 75\%-male training data may give the best average accuracy. The 75\%-female training data yields the smallest gender gap in accuracy, and implies that a gender ratio may be chosen to aim for approximately equal test accuracy - though this accuracy will not be the highest for males, females, or on average. Ultimately, they find no empirical support for the speculation that gender-balanced training data equalizes test accuracy.
\newpage

%% file: tables/ratioTable.tex
\begin{table}[h]
    \centering
    \small
    \begin{tabular}{l|rr|rr}
        \textbf{}& \multicolumn{2}{c|}{\textbf{\# Subjects}}& \multicolumn{2}{c}{\textbf{\# Images}} \\
        \textbf{Subset Name} & \multicolumn{1}{c}{\textbf{Males}} & \multicolumn{1}{c|}{\textbf{Females}} & \multicolumn{1}{c}{\textbf{Males}} & \multicolumn{1}{c}{\textbf{Females}} \\ \hline
        Full & 5,154& 3,477 & 1,828,987& 1,291,873\\
        Balanced & 3,477& 3,477 & 1,291,873& 1,291,873\\ \hline
        F100 & 0& 3,477 & 0& 1,291,873\\
        M25F75 & 870& 2,607 & 322,969& 968,904\\
        M50F50 & 1,739& 1,739 & 645,937& 645,937\\
        M75F25 & 2,607& 870 & 968,904& 322,969\\
        M100 & 3,477& 0 & 1,291,873& 0 \\
        \hline \hline
        Full & 59,563 & 22,499 & 3,741,274 & 1,890,773\\
        Balanced & 22,499 & 22,499 & 1,890,773 & 1,890,773\\ \hline
        F100 & 0 & 22,499 & 0 & 1,890,773\\
        M25F75 & 5,624 & 16,875 & 472,693 & 1,418,080\\
        M50F50 & 11,249 & 11,249 & 945,386 & 945,386\\
        M75F25 & 16,875 & 5,624 & 1,418,080 & 472,693\\
        M100 & 22,499 & 0 & 1,890,773 & 0 
    \end{tabular}
    \vspace{-0.5em}
    \caption{VGGFace2 (top half) and MS1MV2 (bottom half) training subsets created. (Table adapted from \cite{vitor_gender}.)}
    \label{ratioTable}
\end{table}

%% file: tables/full_balanced.tex
\begin{table}
  \centering %
  \caption{Gender accuracy (\%) with TAR@FAR=$0.001\%$ when trained using the entire training dataset (full) and the gender balanced version (balanced). (Table adapted from \cite{vitor_gender}.)}\label{full_balanced}
  \includegraphics[width=\textwidth]{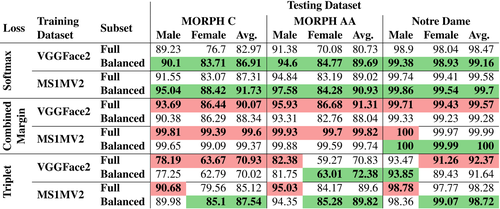}
\end{table}

%% file: Sections/BalancedPixels.tex
\section{Does Balancing Pixels in Test Data Equalize Genuine Distributions?}
\label{sec:balancedpixels}

In \emph{Gendered Differences in Face Recognition
Accuracy Explained by Hairstyles, Makeup, and
Facial Morphology} \cite{vitorTIFS}, Albiero et al. honed in on another speculated cause of the gender gap: a disparity in the number of ``face pixels'' between male and female images. Face images are typically pre-processed with detection, alignment and cropping. Face pixels compose the resultant segment of the image that is directly evaluated by the matcher. The initial experiment analyzes how the quantity of face pixels varies by gender and relates to gender-specific accuracy. A natural follow-up experiment examines recognition accuracy when a dataset is controlled for equal face pixels across genders. 

\subsection{Assessing Impact of Pixels of the Face}
\begin{figure}[!ht]
\floatbox[{\capbeside\thisfloatsetup{capbesideposition={right,center},capbesidewidth=4cm}}]{figure}[\FBwidth]
{\caption{Example images and their BiSeNet-generated binary masks, where white indicates \emph{face} and black \emph{non-face} regions. Note the significant impact of hairstyle on the size of the \emph{face} region - and subsequently, the number of face pixels.}\label{fig:bisenetmasks}}
{\includegraphics[width=6cm]{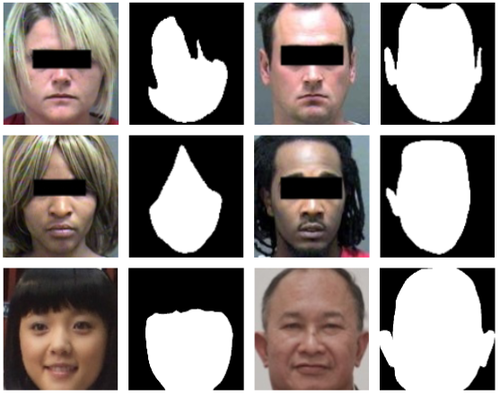}}
\end{figure}

In order to discern between face and non-face pixels, Albiero et al. used Bilateral Segmentation Network (``BiSeNet'') \cite{bisenet} to segment each face and generate a corresponding binary mask. Segmented regions classified as skin, eyebrows, eyes, ears, nose and mouth are considered ``face pixels'' in the mask; regions like neck and hair are excluded. Sample masks are shown in Figure \ref{fig:bisenetmasks}, adapted from \cite{vitorTIFS}. 

\begin{figure}[!ht]
\centering
\includegraphics[width=\columnwidth]{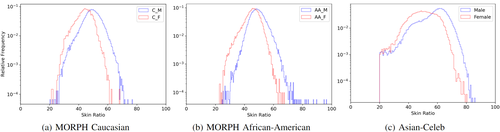}
\caption{Comparison of female and male distributions of percent of image labelled \emph{face}. (Figure adapted from \cite{vitorTIFS}.)}
\label{skinRatios}
\end{figure}

The distribution of the fraction of images labeled \emph{face} is provided in Figure \ref{skinRatios}. For the majority of female faces, 25\% to 45\% of the image represents face for MORPH, and 20\% to 55\% for Asian-Celeb. For the majority of male images, 45\% to 70\% of the image represents face for MORPH, and 55\% to 80\% for Asian-Celeb. The gender-shifted distributions make it clear that, on average, female images contain less usable information about the face. 

\begin{figure}[!ht]
\centering
\includegraphics[width=0.7\columnwidth]{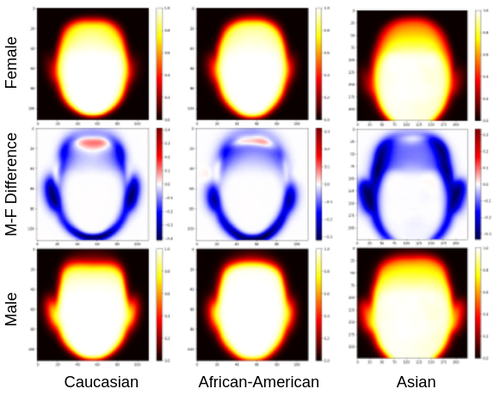}
\caption{Heatmaps representing frequency of pixels labelled \emph{face}. (Figure adapted from \cite{vitorTIFS}.)}
\label{fig:genderheatmaps}
\end{figure}

With the segmented masks, Albiero et al. computed heatmaps to reflect the fraction of images for which a given pixel is labelled \emph{face}. Figure \ref{fig:genderheatmaps} shows the frequency of \emph{face}-labelled pixels for each race and gender. The heatmaps in the middle row of Figure \ref{fig:genderheatmaps} give the male-female difference (``M-F Difference'') for each group. Blue pixels are those more frequently labeled \emph{face} for males than females, with red showing the opposite. In the male heatmaps, the chin, ear, and jawline regions are more prominent than in the female heatmaps, as evidenced by the larger regions of blue pixels in the difference heatmaps. Females appear to have slightly higher foreheads, as indicated by the red regions in the Caucasian and African American difference heatmaps  

Overall, gendered differences in facial morphology and choice of hairstyle result in fewer average face pixels for female subjects. Females already tend to have smaller heads and facial regions, and their hairstyles ``more frequently occlude the ears and part of the sides of the face.'' \cite{vitorTIFS} As with the percent-face distributions (Figure \ref{skinRatios}), Albiero et al. observed that the average face image of a biological female contains a smaller fraction of usable biometric information than that of a biological male. 

\subsection{Equalizing Pixels of the Face}

\begin{figure}[!ht]
\centering
\includegraphics[width=\columnwidth]{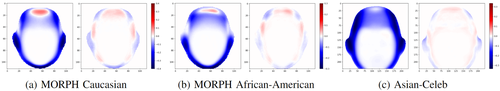}
\caption{The original image set is first masked at 10\% level from female heatmap, giving the left difference heatmaps. A subset of the male images is selected using the intersection-over-union (IoU) to balance with the female images, giving the right difference heatmap. (Figure adapted from \cite{vitorTIFS}.)}
\label{iouMaps}
\end{figure}

Led by the previous finding, Albiero et al. next created a version of the dataset with roughly equal face information. First, to establish the same maximum set of pixels containing face information for males and female, they masked the area outside of the 10\% level in the female heatmap in all images. The 10\% masks are shown on the left in Figure \ref{iouMaps}. As the female face region is generally a subset of the male, this step preserves the majority of available face information. To decrease the remaining bias toward greater information for males, for every female image, they selected a male image with the maximum intersection-over-union (IoU) of pixels labelled \emph{face}. The final heatmaps are given on the right in Figure \ref{iouMaps}. In the resulting ``information equalized'' dataset, the male-female difference in information is minimized. At any pixel, remaining differences are generally less than 5\% and roughly balanced between the two genders.

\begin{figure}[!ht]
\centering
\includegraphics[width=\columnwidth]{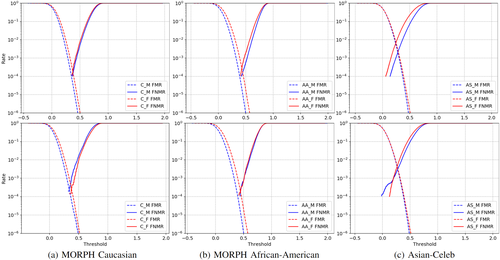}
\caption{Impostor and genuine distributions for original (top) and information-equalized (bottom) female and male image sets. (Figure adapted from \cite{vitorTIFS}.)}
\label{iouIMG}
\end{figure}

The impostor and genuine distributions from the original and information-equalized datasets are given in Figure \ref{iouIMG}. The equalizing step proves highly impactful: though FMR remains worse for females than males, the female FNMR becomes \emph{the same or slightly better} than the male FNMR for African-American and Caucasian groups. While Asian-Celeb features the largest gender gap in FNMR, the equalized subset still pushes the male and female genuine distributions significantly closer.

Notably, these results were solely derived from balancing face pixels in the information-equalized dataset. Both experiments used ArcFace trained on a known female-underrepresented dataset. Albiero et. al conclude that the common observation that the female genuine distribution is worse than male is rooted in gendered social conventions for the hairstyles in test images rather than imbalanced training data. When image datasets used to compute test accuracy are controlled for face-pixel information, the female genuine distribution is as good as (or better than!) the male genuine distribution.

Despite promising results for FNMR, Albiero et al. sought to understand why female FMR remains worse. Even in the information-equalized dataset, on average, a non-mated female pair yields a higher similarity score than a non-mated male pair. They analyzed male and female impostors by performing a clustering experiment using each subject's highest face-pixel image. Each image's ArcFace feature vector was generated, and each vector began as its own cluster. Within the same gender, clusters were combined incrementally using Euclidean distance until only one cluster remained. 

\begin{figure}[!ht]
\centering
\includegraphics[width=\columnwidth]{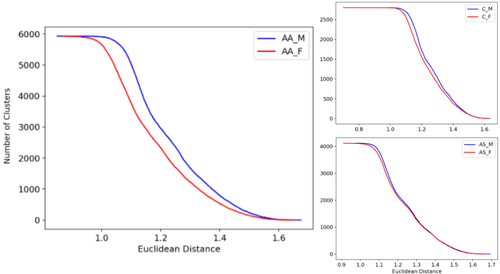}
\caption{Number of clusters as distance threshold increases for ArcFace features. Datasets compared have the same number of subjects, and one image per subject, thus clusters are formed by different persons. (Figure adapted from \cite{vitorTIFS}.)}
\label{clusters}
\end{figure}

The male and female clustering curves for each dataset are shown in Figure \ref{clusters}. The African American male versus female curves, which feature the greatest separation, are enlarged on the left. The right column features the gender-separated curves for Caucasian (top) and Asian (bottom) groups.

For all three datasets, male clusters form at higher thresholds than female, and the male-female difference in the cluster curves is proportional to the difference in FMR. For MORPH African-American, with a large male-female FMR difference, females form clusters at much lower distances than males. For MORPH Caucasian, with a smaller male-female FMR difference, the difference in the cluster curve is smaller. For Asian-Celeb, where the FMR is almost the same for males and females, the number of clusters is also similar. These results imply that images of two different females are more similar than images of two different males, causing females to be clustered together at faster rate, which translates to a worse FMR.

\subsection{Summary of Balanced Pixel Results}

In previous results, females generally have a worse genuine distribution and correspondingly higher FNMR at the same decision threshold as males. However, females also tend to have different hairstyles and face shapes than males, and these factors result in a smaller fraction of an image, on average, containing face information for females. When datasets are controlled to equalize the fraction of the image that represents face, the female genuine distribution is as good as, or better than, the male genuine distribution.  The female impostor distribution remains worse, and is explained by the finding that non-mated female pairs are more similar than non-mated male pairs.

Thus, the result that females have a higher FNMR than males is found to be caused by gendered differences in facial appearance. Because hairstyle conventions and face morphology may vary between racial groups, this conclusion is consistent with, and may explain, the observation in the NIST report \cite{nist_demographics} that the gender gap in FNMR is not universal across races.

\newpage

%% file: Sections/Discussion.tex
\section{Discussion}
\label{sec:discussion}

In recent years, public discourse around race and gender equity has piqued great interest in the subject of face recognition. Research studies have consistently reported that women and darker-skinned individuals are more likely to be falsely accepted or rejected by recognition systems. These demographic-specific accuracy gaps are quantifiable as differences in both false match and non-match rates and impostor and genuine score distributions. However, the \textit{causes} of the observed differences are not so readily apparent. 

The previous sections explore four commonly speculated causes. Each experiment is designed to isolate a single factor and remove confounding variables in analysis of corresponding recognition accuracy. Evidence is not found to support the idea that skin tone or face morphology alone decrease performance, or that balancing training data increases performance. Ultimately, controlling for equal cross-gender face pixel information is the only factor found to notably improve the overall female FNMR. However, the gender gap persists in FMR, and the finding that female faces are inherently more similar than male faces suggests that the issue requires further study. 

In practice, test data is generally not evaluated for ``fairness''. That is, research often simply assumes that the test data is free of any factors that would bias the accuracy results. Then any demographic difference in accuracy is attributed to algorithms or training data, rather than inherent demographic-related differences in the test data. As the balanced-pixel experiment reveals, gendered social conventions and aesthetic choices like hairstyle can cause ``bias'' toward less information available for face recognition for females. These and other image quality issues differ across demographic groups, and, when unaccounted for, manifest as a gender gap in recognition accuracy.

In an increasingly digital world, automated technologies like face recognition have direct, and significant, impact on all people. As of yet, there is no complete or definitive answer to \textit{why} face recognition performance varies across race and gender groups. But we must persist in noting these disparities and assessing them experimentally. Such research is essential to ensuring the very purpose of the technology: to better the lives of its users.
\clearpage
 